\documentclass[11pt,a4paper]{article}
\usepackage[acceptedWithA]{tacl2018v2}
\usepackage{times,latexsym}
\usepackage{url}
\usepackage[T1]{fontenc}
\usepackage[titletoc,title]{appendix}

\usepackage{latexsym}
\usepackage{mathtools}
\usepackage{graphicx}
\usepackage{wrapfig}
\usepackage{amsmath}
\usepackage{subcaption}
\usepackage{balance}
\usepackage{enumitem}
\usepackage{float}
\usepackage{framed} 
\usepackage{verbatim}

\usepackage{xspace}
\usepackage{xargs} 
\usepackage[colorinlistoftodos,prependcaption,textsize=tiny]{todonotes}
\usepackage{marginnote}
\usepackage{color}
\definecolor{darkgreen}{RGB}{0,100,0}

\usepackage{cleveref}
\Crefformat{equation}{(#2#1#3)}

\usepackage{microtype}
\usepackage{booktabs}
\usepackage{rotating}
\usepackage{changepage}

\usepackage{pifont} 
\newcommand{\cmark}{\ding{51}}%
\newcommand{\xmark}{\ding{55}}%

\newcommandx{\icmtr}[2][1=]{\todo[inline]{SR: #2}\xspace}
\newcommandx{\icmtl}[2][1=]{\todo[inline]{DC: #2}\xspace}
\newcommandx{\icmtm}[2][1=]{\todo[inline]{CM: #2}\xspace}

\newcommandx{\cmtr}[2][1=]{\todo[linecolor=red,backgroundcolor=red!25,bordercolor=red,#1]{SR: #2}\xspace}
\newcommandx{\cmtl}[2][1=]{\todo[linecolor=yellow,backgroundcolor=red!25,bordercolor=red,#1]{DC: #2}\xspace}
\newcommandx{\cmtm}[2][1=]{\todo[linecolor=blue,backgroundcolor=blue!10,bordercolor=blue,#1]{CM: #2}\xspace}
\newcommand{\ignore}[1]{}

\definecolor{CMpurple}{rgb}{0.6,0.18,0.64}

\title{CoQA: A Conversational Question Answering Challenge}

\author{Siva Reddy$^*$ \quad Danqi Chen$^*$ \quad Christopher D. Manning \\
  Computer Science Department \\
  Stanford University \\
  {\tt \{sivar,danqi,manning\}@cs.stanford.edu}
}

\date{}

\begin{document}
\maketitle

\renewcommand{\thefootnote}{\fnsymbol{footnote}}
\footnotetext[1]{The first two authors contributed equally.}
\renewcommand{\thefootnote}{\arabic{footnote}}

\begin{abstract}
Humans gather information through conversations involving a series of interconnected questions and answers.
For machines to assist in information gathering, it is therefore essential to enable them to answer conversational questions.
We introduce CoQA, a novel dataset for building \textbf{Co}nversational \textbf{Q}uestion \textbf{A}nswering systems.\footnote{CoQA is pronounced as \textit{coca}.}
Our dataset contains 127k questions with answers, obtained from 8k conversations about text passages from seven diverse domains.
The questions are conversational, and the answers are free-form text with their corresponding evidence highlighted in the passage.
We analyze CoQA in depth and show that conversational questions have challenging phenomena not present in existing reading comprehension datasets, e.g., coreference and pragmatic reasoning.
We evaluate strong dialogue and reading comprehension models on CoQA.
The best system obtains an F1 score of 65.4\%, which is 23.4~points behind human performance (88.8\%), indicating there is ample room for improvement.
We present CoQA as a challenge to the community at \url{https://stanfordnlp.github.io/coqa}.

\end{abstract}


\section{Introduction}
\label{sec:intro}
%

%


We ask other people a question to either seek or test their knowledge about a subject.
Depending on their answer, we follow up with another question and their second answer builds on what has already been discussed.
This incremental aspect makes human conversations succinct.
An inability to build and maintain common ground in this way is part of why virtual assistants usually don't seem like competent conversational partners.
In this paper, we introduce CoQA, a \textbf{Co}nversational \textbf{Q}uestion \textbf{A}nswering dataset for measuring the ability of machines to participate in a question-answering style conversation.
In CoQA, a machine has to understand a text passage and answer a series of questions that appear in a conversation.
We develop CoQA with three main goals in mind.


\begin{figure}[tp]
\footnotesize
\begin{tabular}{p{\columnwidth}}
\midrule
Jessica went to sit in her rocking chair. Today was her birthday and she was turning 80. Her granddaughter Annie was coming over in the afternoon and Jessica was very excited to see her. Her daughter Melanie and Melanie's husband Josh were coming as well. Jessica had $\ldots$\\
\\
Q$_1$:		Who had a birthday? \\
A$_1$: Jessica \\
R$_1$: Jessica went to sit in her rocking chair. Today was her birthday and she was turning 80.\\
\vspace{0em}
Q$_2$: How old would she be?\\
A$_2$: 80 \\
R$_2$: she was turning 80 \\
\vspace{0em}
Q$_3$: Did she plan to have any visitors?\\
A$_3$: Yes \\
R$_3$: Her granddaughter Annie was coming over \\
\vspace{0em}
Q$_4$: How many?\\
A$_4$: Three \\
R$_4$: Her granddaughter Annie was coming over in the afternoon and Jessica was very excited to see her. Her daughter Melanie and Melanie's husband Josh were coming as well. \\
\vspace{0em}
Q$_5$: Who?\\
A$_5$: Annie, Melanie and Josh \\
R$_5$: Her granddaughter Annie was coming over in the afternoon and Jessica was very excited to see her. Her daughter Melanie and Melanie's husband Josh were coming as well.\\
\bottomrule
\end{tabular}
\caption{A conversation from the CoQA dataset. Each turn contains a question (Q$_i$), an answer (A$_i$) and a rationale (R$_i$) that supports the answer.}
\label{fig:question-types}
\end{figure}

\begin{table*}[tp]
\footnotesize
\hspace{-1em}\begin{tabular}{lcll}
\midrule
{Dataset} & {Conversational} & {Answer Type} & {Domain} \\
\midrule
MCTest \cite{richardson_mctest_2013} & \xmark &  Multiple choice & Children's stories \\
CNN/Daily Mail  \cite{hermann_teaching_2015} & \xmark & Spans  & News \\
Children's book test  \cite{hill_goldilocks_2016} & \xmark &  Multiple choice  & Children's stories \\
SQuAD \cite{rajpurkar_squad_2016}  & \xmark  &  Spans  & Wikipedia \\
MS MARCO  \cite{nguyen_ms_2016} & \xmark &  Free-form text, Unanswerable & Web Search \\
NewsQA \cite{trischler_newsqa_2017}  & \xmark & Spans  & News  \\
SearchQA \cite{dunn_searchqa_2017} & \xmark & Spans & Jeopardy \\
TriviaQA \cite{joshi_triviaqa_2017} & \xmark & Spans  & Trivia \\
RACE \cite{lai_race_2017}  & \xmark & Multiple choice &  Mid/High School Exams\\
Narrative QA \cite{kocisky_narrativeqa_2018} & \xmark & Free-form text & Movie Scripts, Literature \\
SQuAD 2.0 \cite{rajpurkar_know_2018}  & \xmark  &  Spans, Unanswerable  & Wikipedia \\
\midrule
CoQA (this work) & \cmark & Free-form text, Unanswerable; &  Children's Stories, Literature, \\
 & &  Each answer comes with a  &  Mid/High School Exams, News, \\
& &   text span rationale & Wikipedia, Reddit, Science\\
\midrule
\end{tabular}
\caption{Comparison of CoQA with existing reading comprehension datasets.}
\label{tab:datasets}
\end{table*}

The first concerns the nature of questions in a human conversation.
\Cref{fig:question-types} shows a conversation between two humans who are reading a passage, one acting as a questioner and the other as an answerer.
In this conversation, every question after the first is dependent on the conversation history.
For instance, Q$_5$ (\textit{Who?}) is only a single word and is impossible to answer without knowing what has already been said.
Posing short questions is an effective human conversation strategy, but such questions are really difficult for machines to parse. As is well known, state-of-the-art models rely heavily on lexical similarity between a question and a passage \cite{chen_thorough_2016,weissenborn_making_2017}.
At present, there are no large-scale reading comprehension datasets which contain questions that depend on a conversation history (see \Cref{tab:datasets}) and this is what CoQA is mainly developed for.\footnote{Concurrent with our work, \newcite{choi_quac_2018} also created a conversational dataset with a similar goal, but it differs in many aspects. We discuss the details in \Cref{sec:related-work}.}
The second goal of CoQA is to ensure the naturalness of answers in a conversation.
Many existing QA datasets restrict answers to contiguous text spans in a given passage (\Cref{tab:datasets}).
Such answers are not always natural, for example, there is no span-based answer to Q$_4$ (\textit{How many?}) in \Cref{fig:question-types}. In CoQA, we propose that the answers can be free-form text, while for each answer, we also provide a text span from the passage as a rationale to the answer. Therefore, the answer to Q$_4$ is simply \textit{Three} while its rationale spans across multiple sentences. Free-form answers have been studied in previous reading comprehension datasets e.g., MS MARCO \cite{nguyen_ms_2016} and NarrativeQA \cite{kocisky_narrativeqa_2018} and metrics such as BLEU or ROUGE are used for evaluation due to the high variance of possible answers. One key difference in our setting is that we require answerers to first select a text span as the rationale and then edit it to obtain a free-form answer.\footnote{In contrast, in NarrativeQA, the annotators were encouraged to use their own words and copying was not allowed in their interface.} Our method strikes a balance between naturalness of answers and reliable automatic evaluation, and it results in a high human agreement (88.8\% F1 word overlap among human annotators).
The third goal of CoQA is to enable building QA systems that perform robustly across domains.
The current QA datasets mainly focus on a single domain which makes it hard to test the generalization ability of existing models.
Hence we collect our dataset from seven different domains --- children's stories, literature, middle and high school English exams, news, Wikipedia, Reddit and science.
The last two are used for out-of-domain evaluation.

To summarize, CoQA has the following key characteristics:

\begin{itemize}
\item It consists of 127k conversation turns collected from 8k conversations over text passages. The average conversation length is 15 turns, and each turn consists of a question and an answer.
\item It contains free-form answers and each answer has a span-based rationale highlighted in the passage.
\item Its text passages are collected from seven diverse domains: five are used for in-domain evaluation and two are used for out-of-domain evaluation.
\end{itemize}

Almost half of CoQA questions refer back to conversational history using anaphors, and a large portion require pragmatic reasoning making it challenging for models that rely on lexical cues alone.
We benchmark several deep neural network models, building on top of state-of-the-art conversational and reading comprehension models (\Cref{sec:models}).
The best-performing system achieves an F1 score of 65.4\%. In contrast, humans achieve 88.8\% F1, 23.4\% F1 higher, indicating that there is a lot of headroom for improvement.




\section{Task Definition}
Given a passage and a conversation so far, the task is to answer the next question in the conversation.
Each turn in the conversation contains a question and an answer.

For the example in \Cref{fig:coreference}, the conversation begins with question Q$_1$.
We answer Q$_1$ with A$_1$ based on the evidence R$_1$, which is a contiguous text span from the passage.
In this example, the answerer only wrote the \textit{Governor} as the answer but selected a longer rationale \textit{The Virginia governor's race}.

When we come to Q$_2$ (\textit{Where?}), we must refer back to the conversation history otherwise its answer could be \textit{Virginia} or \textit{Richmond} or something else.
In our task, conversation history is indispensable for answering many questions.
We use conversation history Q$_1$ and A$_1$ to answer Q$_2$ with A$_2$ based on the evidence R$_2$.
Formally, to answer Q$_n$, it depends on the conversation history: Q$_1$, A$_1$, $\ldots$, Q$_{n-1}$, A$_{n-1}$.
For an unanswerable question, we give \textit{unknown} as the final answer and do not highlight any rationale.

In this example, we observe that the entity of focus changes as the conversation progresses.
The questioner uses \textit{his} to refer to \textit{Terry} in Q$_4$ and \textit{he} to \textit{Ken} in Q$_5$.
If these are not resolved correctly, we end up with incorrect answers.
The conversational nature of questions requires us to reason from multiple sentences (the current question and the previous questions or answers, and sentences from the passage).
It is common that a single question may require a rationale spanning across multiple sentences (e.g., Q$_1$ Q$_4$ and Q$_5$ in \Cref{fig:question-types}).
We describe additional question and answer types in \Cref{sec:analysis}.

Note that we collect rationales as (optional) evidence to help answer questions. However, they are not provided at testing time. A model needs to decide on the evidence by itself and derive the final answer.

\begin{figure}[t]
\footnotesize
\hspace{-1em}\begin{tabular}{p{\columnwidth}}
\midrule
The Virginia governor's race, billed as the marquee battle of an otherwise anticlimactic 2013 election cycle, is shaping up to be a foregone conclusion. Democrat Terry McAuliffe, the longtime political fixer and moneyman, hasn't trailed in a poll since May. Barring a political miracle, Republican Ken Cuccinelli will be delivering a concession speech on Tuesday evening in Richmond. In recent ...\\
\\
Q$_1$:               What are the candidates {\bf \color{blue} running} for?\\
A$_1$:               Governor\\
R$_1$: The Virginia governor's race\\
\vspace{0em}
Q$_2$:               {\bf \color{blue} Where}?\\
A$_2$:               Virginia \\
R$_2$: The Virginia governor's race\\
\vspace{0em}
Q$_3$:               Who is the democratic candidate?\\
\vspace{-0.6em}{\bf \color{orange} A$_3$}:               {\bf \color{darkgreen} Terry McAuliffe} \\
R$_3$: Democrat Terry McAuliffe\\
\vspace{0em}
Q$_4$:               Who is {\bf \color{darkgreen} his} opponent?\\
\vspace{-0.6em}{\bf \color{orange} A$_4$}:               {\bf \color{red} Ken Cuccinelli} \\
R$_4$ Republican Ken Cuccinelli\\
\vspace{0em}
Q$_5$:               What party does {\bf \color{red} he} belong to?\\
A$_5$:               Republican \\
R$_5$: Republican Ken Cuccinelli\\
\vspace{0em}
Q$_6$:               Which of {\bf \color{orange} them} is winning?\\
A$_6$:               Terry McAuliffe \\
R$_6$: Democrat Terry McAuliffe, the longtime political fixer and moneyman, hasn't trailed in a poll since May\\
\bottomrule
\end{tabular}
\caption{A conversation showing coreference chains in color. The entity of focus changes in Q4, Q5, Q6.}
\label{fig:coreference}
\end{figure}


\section{Dataset Collection}
For each conversation, we employ two annotators, a questioner and an answerer.
This setup has several advantages over using a single annotator to act both as a questioner and an answerer:
1) when two annotators chat about a passage, their dialogue flow is natural;
2) when one annotator responds with a vague question or an incorrect answer, the other can raise a flag which we use to identify bad workers;
and 3) the two annotators can discuss guidelines (through a separate chat window) when they have disagreements.
These measures help to prevent spam and to obtain high agreement data.\footnote{Due to AMT terms of service, we allowed a single worker to act both as a questioner and an answerer after a minute of waiting.
This constitutes around 12\% of the data.
We include this data in the training set only. }
We use Amazon Mechanical Turk (AMT) to pair workers on a passage through the ParlAI MTurk API \cite{miller_parlai_2017}.


\subsection{Collection Interface}
\label{sec:interface}
We have different interfaces for a questioner and an answerer (see Appendix).
A questioner's role is to ask questions, and an answerer's role is to answer questions in addition to highlighting rationales.
Both questioner and answerer sees the conversation that happened until now, i.e., questions and answers from previous turns and rationales are kept hidden.
While framing a new question, we want questioners to avoid using exact words in the passage in order to increase lexical diversity.
When they type a word that is already present in the passage, we alert them to paraphrase the question if possible.
While answering, we want answerers to stick to the vocabulary in the passage in order to limit the number of possible answers.
We encourage this by asking them to first highlight a rationale (text span), which is then automatically copied into the answer box, and we further ask them to edit the copied text to generate a natural answer. We found 78\% of the answers have at least one edit such as changing a word's case or adding a punctuation.

\subsection{Passage Selection}
We select passages from seven diverse domains: children's stories from MCTest \cite{richardson_mctest_2013}, literature from Project Gutenberg\footnote{Project Gutenberg \url{https://www.gutenberg.org}}, middle and high school English exams from RACE \cite{lai_race_2017}, news articles from CNN \cite{hermann_teaching_2015}, articles from Wikipedia, Reddit articles from the Writing Prompts dataset \cite{fan_hierarchical_2018} and science articles from AI2 Science Questions \cite{welbl_crowdsourcing_2017}.

Not all passages in these domains are equally good for generating interesting conversations.
A passage with just one entity often results in questions that entirely focus on that entity.
Therefore, we select passages with multiple entities, events and pronominal references  using Stanford CoreNLP \cite{manning_stanford_2014}.
We truncate long articles to the first few paragraphs that result in around 200 words.

\Cref{tab:coqa-domain-distribution} shows the distribution of domains.
We reserve the Reddit and Science domains for out-of-domain evaluation.
For each in-domain dataset, we split the data such that there are 100 passages in the development set, 100 passages in the test set, and the rest in the training set.
For each out-of-domain dataset, we only have 100 passages in the test set.

\begin{table}
\footnotesize
\hspace{-1em}\begin{tabular}{p{1.9cm}rrrr}
\toprule
Domain & \#Passages & \#Q/A & Passage    & \#Turns per \\
 & & pairs & length & passage \\
\midrule
\multicolumn{5}{c}{In-domain} \\
\midrule
Children's Sto.  & 750 & 10.5k & 211 &  14.0 \\
Literature  & 1,815 & 25.5k & 284  & 15.6 \\
Mid/High Sch. & 1,911 & 28.6k & 306  & 15.0 \\
News & 1,902 & 28.7k & 268 &  15.1 \\
Wikipedia & 1,821 & 28.0k & 245  & 15.4 \\
\midrule
\multicolumn{5}{c}{Out-of-domain} \\
\midrule
Reddit & 100 & 1.7k & 361 & 16.6 \\
Science & 100 & 1.5k & 251  & 15.3\\
\midrule
Total & 8,399 & 127k  & 271 & 15.2 \\
\bottomrule
\end{tabular}
\caption{Distribution of domains in CoQA.}
\label{tab:coqa-domain-distribution}
\end{table}

\begin{figure*}
\begin{subfigure}{0.45\textwidth}
\includegraphics[trim=22em 55em 193em 22em,clip=true,width=1.05\columnwidth]{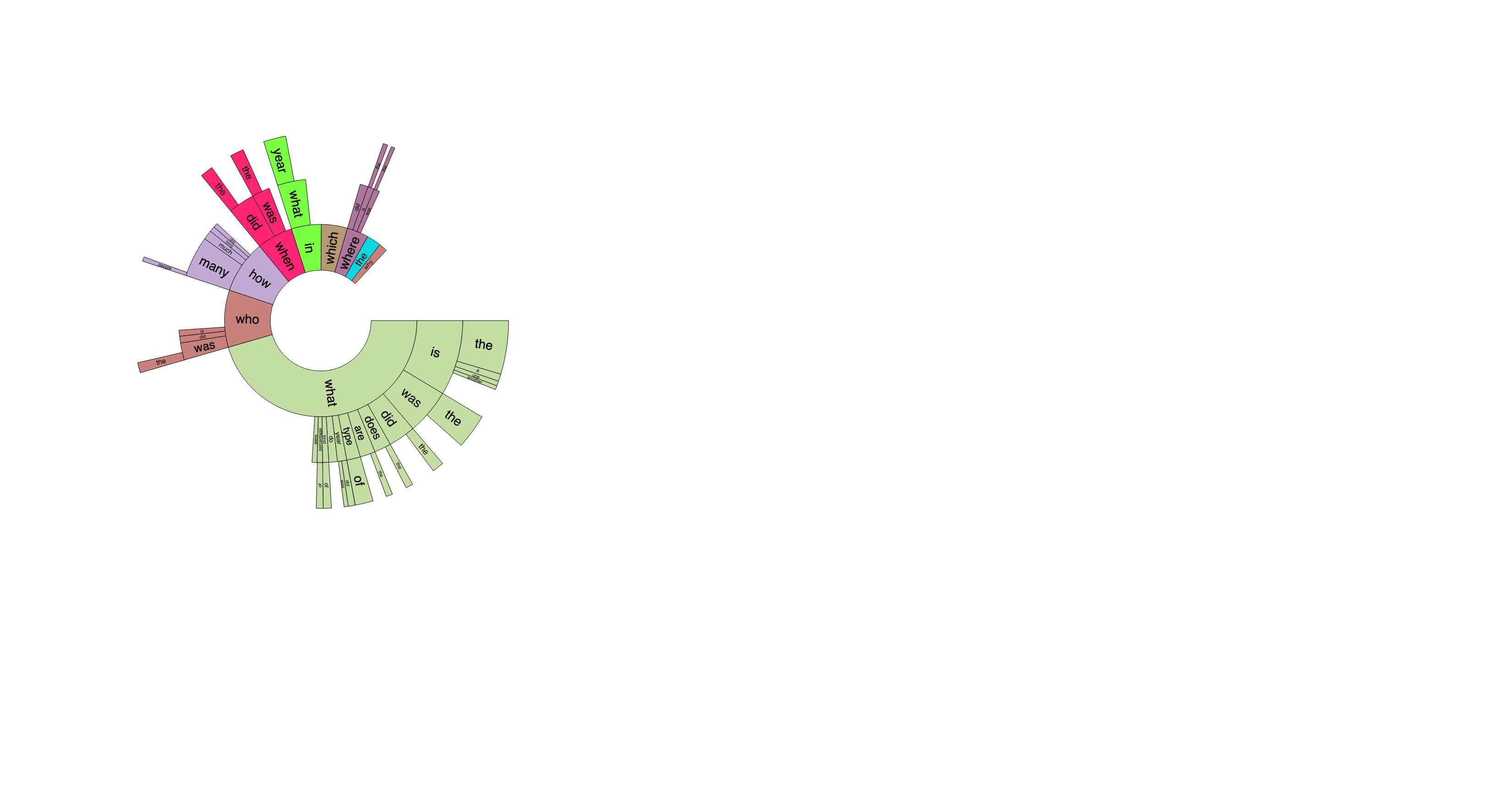}
\caption{SQuAD}
\label{fig:squad-questions}
\end{subfigure}
\hspace{0.5em}
\begin{subfigure}{0.45\textwidth}
\vspace{-1em}
\includegraphics[trim=22em 55em 197em 22em,clip=true,width=1.05\columnwidth]{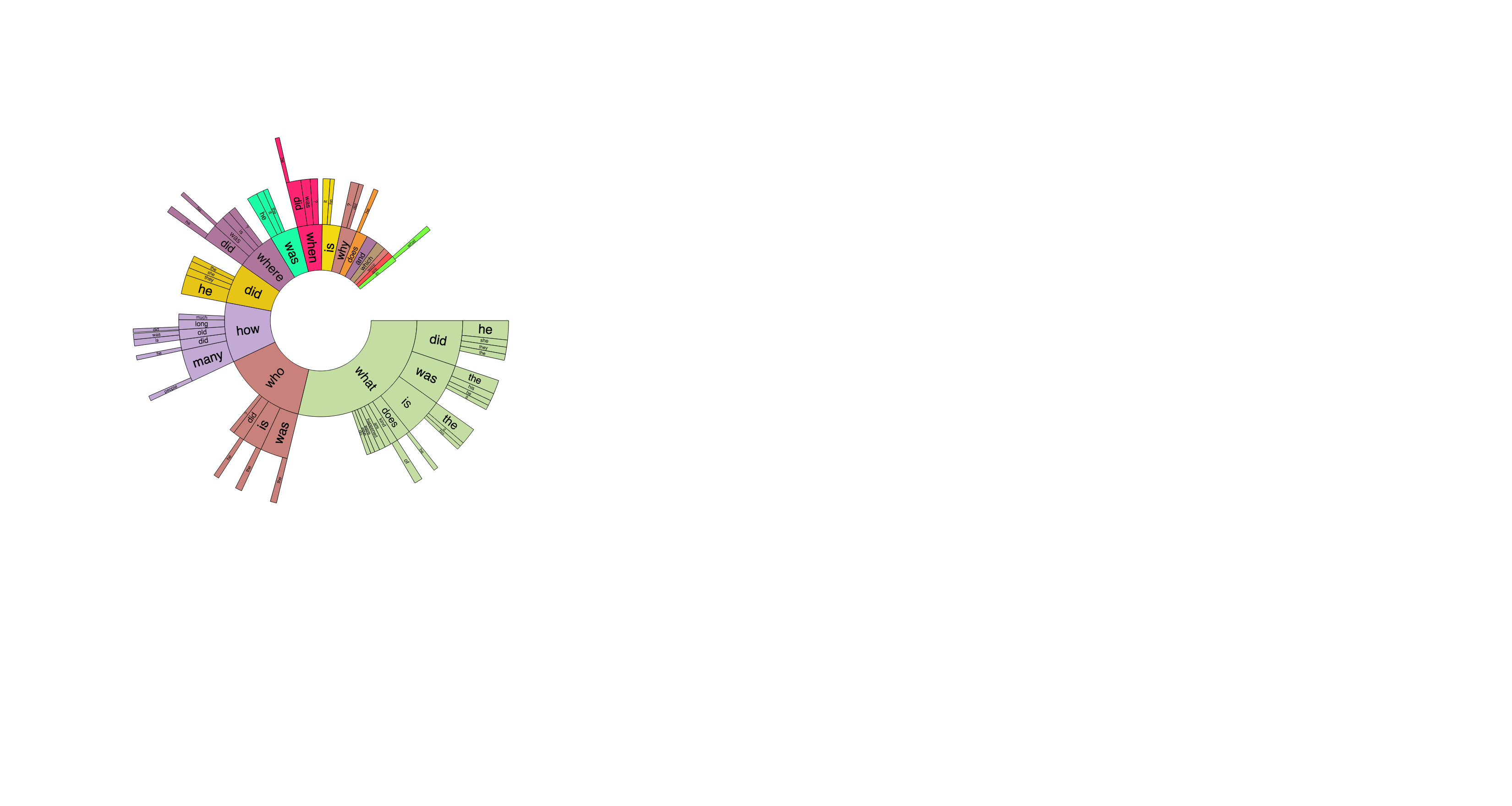}
\caption{CoQA}
\label{fig:coqa-questions}
\end{subfigure}
\caption{Distribution of trigram prefixes of questions in SQuAD and CoQA.}
\label{fig:squad-coqa-comp}
\end{figure*}

\subsection{Collecting Multiple Answers}
\label{sec:multiple-answers}
Some questions in CoQA may have multiple valid answers.
For example, another answer to Q$_4$ in \Cref{fig:coreference} is \textit{A Republican candidate}.
In order to account for answer variations, we collect three additional answers for all questions in the development and test data.
Since our data is conversational, questions influence answers which in turn influence the follow-up questions.
In the previous example, if the original answer was \textit{A Republican Candidate}, then the following question \textit{Which party does he belong to?}\ would not have occurred in the first place.
When we show questions from an existing conversation to new answerers, it is likely they will deviate from the original answers which makes the conversation incoherent.
It is thus important to bring them to a common ground with the original answer.

We achieve this by turning the answer collection task into a game of predicting original answers.
First, we show a question to an answerer, and when she answers it, we show the original answer and ask her to verify if her answer matches the original.
For the next question, we ask her to guess the original answer and verify again.
We repeat this process with the same answerer until the conversation is complete.
The entire conversation history is shown at each turn (question, answer, original answer for all previous turns but not the rationales).
In our pilot experiment, the human F1 score is increased by 5.4\% when we use this verification setup.

\begin{table*}[t]
\footnotesize
\begin{minipage}{0.3\textwidth}
\begin{minipage}{\textwidth}
\begin{tabular}{p{2.1cm} r r}
\toprule
 & SQuAD  & CoQA \\
\midrule
Passage Length & 117 & 271 \\
Question Length & 10.1 & 5.5 \\
Answer  Length & 3.2 & 2.7 \\
\midrule
\end{tabular}
\vspace{-1em}
\caption{Average number of words in passage, question and answer in SQuAD and CoQA.}
\label{tab:squad-coqa-length}
\end{minipage}\\
\vspace{0.3em}
\begin{minipage}{\textwidth}
\end{minipage}\\
\begin{minipage}{\textwidth}
\begin{tabular}{l r r}
\toprule
& SQuAD   & CoQA  \\
\midrule
Answerable & 66.7\% & 98.7\% \\
Unanswerable & 33.3\% & 1.3\% \\
\midrule
Span found & 100.0\% & 66.8\% \\
No span found & 0.0\% & 33.2\% \\
\midrule
Named Entity & 35.9\% & 28.7\% \\
Noun Phrase & 25.0\% & 19.6\% \\
Yes & 0.0\% & 11.1\% \\
No & 0.1\% & 8.7\% \\
Number & 16.5\% & 9.8\% \\
Date/Time & 7.1\% & 3.9\% \\
Other & 15.5\% & 18.1\% \\
\bottomrule
\end{tabular}
\caption{Distribution of answer types in SQuAD and CoQA.}
\label{tab:squad-coqa-answers}
\end{minipage}
\end{minipage}
\hspace{3em}
\begin{minipage}{0.53\textwidth}
\footnotesize
\vspace{-1.2em}
\begin{tabular}{lp{5.5cm}c}
\toprule
Phenomenon & Example & Percentage \\
\midrule
\multicolumn{3}{c}{Relationship between a question and its passage} \\
\midrule
Lexical match & Q: Who had to rescue her?& 29.8\% \\
& A: the coast guard \\
& R: Outen was rescued by the coast guard \\
Paraphrasing & Q: Did the wild dog \textbf{approach}? & 43.0\% \\
& A: Yes \\
& R: he \textbf{drew} cautiously \textbf{closer} \\
Pragmatics &  Q:               Is Joey a male or female?  &  27.2\% \\
 & A:  Male \\
& R: it looked like a stick man so she kept \textbf{him}. She named her new noodle friend Joey \\
\midrule
\multicolumn{3}{c}{Relationship between a question and its conversation history} \\
\midrule
No coref. & Q: What is IFL? & 30.5\% \\
Explicit coref. & Q: Who had Bashti forgotten? & 49.7\% \\
& A: the puppy \\
& Q: What was \textbf{his} name? \\
Implicit coref. & Q: When will Sirisena be sworn in? & 19.8\% \\
& A: 6 p.m local time  \\
& Q: \textbf{Where}?\\
\bottomrule
\end{tabular}
\caption{Linguistic phenomena in CoQA questions.}
\label{tab:ling-phenomena}
\end{minipage}
\end{table*}


\section{Dataset Analysis}
\label{sec:analysis}

What makes the CoQA dataset conversational compared to existing reading comprehension datasets like SQuAD?
What linguistic phenomena do the questions in CoQA exhibit?
How does the conversation flow from one turn to the next?
We answer these questions below.

\subsection{Comparison with SQuAD 2.0}
SQuAD has been the main benchmark for reading comprehension.  In the following, we perform an in-depth comparison of CoQA and the latest version of SQuAD \cite{rajpurkar_know_2018}.
\Cref{fig:squad-questions} and \Cref{fig:coqa-questions} show the distribution of frequent trigram prefixes.
Because of the free-form nature of answers, we expect a richer variety of questions in CoQA than that in SQuAD.
While nearly half of SQuAD questions are dominated by \textit{what} questions, the distribution of CoQA is spread across multiple question types.
Several sectors indicated by prefixes \textit{did, was, is, does} and \textit{and} are frequent in CoQA but are completely absent in SQuAD. While coreferences are non-existent in SQuAD, almost every sector of CoQA contains coreferences (\textit{he, him, she, it, they})  indicating CoQA is highly conversational.

Since a conversation is spread over multiple turns, we expect conversational questions and answers to be shorter than in a standalone interaction.
In fact, questions in CoQA can be made up of just one or two words (\textit{who?}, \textit{when?},  \textit{why?}).
As seen in \Cref{tab:squad-coqa-length}, on average, a question in CoQA is only 5.5 words long while it is 10.1 for SQuAD.
The answers are a bit shorter in CoQA than SQuAD because of the free-form nature of the answers.

\Cref{tab:squad-coqa-answers} provides insights into the type of answers in SQuAD and CoQA.
While the original version of SQuAD \cite{rajpurkar_squad_2016} does not have any unanswerable questions, the later version \cite{rajpurkar_know_2018} focuses solely on obtaining them resulting in higher frequency than in CoQA.
SQuAD has 100\% span-based answers by design, whereas in CoQA, 66.8\% of the answers overlap with the passage after ignoring punctuation and case mismatches.\footnote{If punctuation and case are not ignored, only 37\% of the answers can be found as spans.}
The rest of the answers, 33.2\%, do not exactly overlap with the passage (see \Cref{sec:abstractive}).
It is worth noting that CoQA has 11.1\% and 8.7\% questions with \textit{yes} or \textit{no} as answers whereas SQuAD has 0\%.
Both datasets have a high number of named entities and noun phrases as answers.

\subsection{Linguistic Phenomena}
\label{sec:ling-phenomena}
We further analyze the questions for their relationship with the passages and the conversation history.
We sample 150 questions in the development set and annotate various phenomena as shown in \Cref{tab:ling-phenomena}.

If a question contains at least one content word that appears in the rationale, we classify it as \textit{lexical match}.
These comprise around 29.8\% of the questions.
If it has no lexical match but is a paraphrase of the rationale, we classify it as \textit{paraphrasing}.
These questions contain phenomena such as synonymy, antonymy, hypernymy, hyponymy and negation.
These constitute a large portion of questions, around 43.0\%.
The rest, 27.2\%, have no lexical cues, and we classify them as \textit{pragmatics}.
These include phenomena like common sense and presupposition.
For example, the question \textit{Was he loud and boisterous?} is not a direct paraphrase of the rationale \textit{he dropped his feet with the lithe softness of a cat} but the rationale combined with world knowledge can answer this question.

For the relationship between a question and its conversation history, we classify questions into whether they are dependent or independent on the conversation history. If dependent, whether the questions contain an explicit marker or not.
Our analysis shows that around 30.5\% questions do not rely on coreference with the conversational history and are answerable on their own. Almost half of the questions (49.7\%) contain explicit coreference markers such as \textit{he, she, it}. These either refer to an entity or an event introduced in the conversation.
The remaining 19.8\% do not have explicit coreference markers but refer to an entity or event implicitly (these are often cases of \emph{ellipsis}, as in the examples in Table~\ref{tab:ling-phenomena}).

\begin{table*}
\footnotesize
\begin{tabular}{l l r}
\toprule
{Answer Type} &  {Example} & {Percentage} \\
\midrule
Yes & Q: is MedlinePlus optimized for mobile? &  48.5\%\\
& A: Yes \\
& R: There is also a site optimized for display on mobile devices \\
No & Q: Is it played outside? &  30.3\% \\
& A: No \\
& R: AFL is the highest level of professional indoor American football \\
Fluency & Q: Why? &  14.3\% \\
& A: so the investigation could continue \\
& R: while the investigation continued \\
Counting &  Q: how many languages is it offered in? & 5.1\% \\
& A: Two \\
& R: The service provides curated consumer health information in English and Spanish \\
Multiple choice  &  Q: Is Jenny older or younger?  & 1.8\%  \\
& A: Older \\
& R: her baby sister is crying so loud that Jenny can't hear herself \\
\midrule
\multicolumn{3}{c}{Fine grained breakdown of Fluency} \\
\midrule
Multiple edits & Q: What did she try just before that? & 41.4\% \\
& A: She \textbf{gave} her \textbf{a} toy \textbf{horse}. \\
& R: She \textbf{would give} her \textbf{baby sister one of her} toy horses. \\
& (morphology: give $\rightarrow$ gave, horses $\rightarrow$ horse; delete: would, baby sister one of her; insert: a) \\
Coreference insertion & Q: what is the cost to end users? & 16.0\%\\
& A: \textbf{It} is free \\
& R: The service is funded by the NLM and \textbf{is free} to users\\
Morphology & Q: Who was messing up the neighborhoods? & 13.9\% \\
& A: vandals \\
& R: \textbf{vandalism} in the neighborhoods\\
Article insertion & Q: What would they cut with? & 	7.2\% \\
& A: \textbf{an} ax \\
& R: the heavy ax \\
Adverb insertion	& Q: How old was the diary? & 4.2\% \\
& A: 190 years \textbf{old} \\
& R: kept 190 years ago \\
Adjective deletion & Q: What type of book? & 4.2\% \\
& A: A diary. \\
& R: a \textbf{120-page} diary \\
Preposition insertion & how long did it take to get to the fire? & 3.4\%\\
& A: \textbf{Until} supper time! \\
& R: By the time they arrived, it was almost supper time. \\
Adverb deletion	& Q: What had happened to the ice? & 3.0\% \\
& A: It had changed \\
& R: It had \textbf{somewhat} changed its formation when they approached it \\
Conjunction insertion	& Q: what else do they get for their work? & 1.3\% \\
& A: potatoes \textbf{and} carrots \\
& R: paid well, both in potatoes, carrots \\
Noun insertion	& Q: Who did & 1.3\%\\
& A: Comedy Central \textbf{employee} \\
& R: But it was a Comedy Central account \\
Coreference deletion	&  Q: What is the story about? &	1.2\% \\
& A: A girl and a dog \\
& R: This is the story of a young girl and \textbf{her} dog \\
Noun deletion	& Q: What is the ranking in the country in terms of people studying? & 	0.8\% \\
& A: the fourth largest population \\
& R: and has the fourth largest \textbf{student} population \\
Possesive insertion	 & Q: Whose diary was it? &	0.8\% \\
& A: Deborah Logan\textbf{'s} \\
& R: a 120-page diary kept 190 years ago by Deborah Logan \\
Article deletion	& Q: why? & 	0.8\% \\
& A: They were going to the circus \\
& R: They \textbf{all} were going to the circus  to see the clowns \\
\bottomrule
\end{tabular}
\caption{Analysis of answers which don't overlap with passage.}
\label{tab:abstractive-answers}
\end{table*}

\subsection{Analysis of Free-form Answers}
\label{sec:abstractive}
Due to the free-form nature of CoQA's answers, around 33.2\% of them do not exactly overlap with the given passage.
We analyze 100~conversations to study the behavior of such answers.\footnote{We only pick the questions in which none of its answers can be found as a span in the passage.}
As shown in \Cref{tab:abstractive-answers}, the answers \textit{Yes} and \textit{No} constitute 48.5\% and 30.3\% respectively, totaling 78.8\%.
The next majority, around 14.3\%, are edits to text spans to improve the fluency (naturalness) of answers.
More than two thirds of these edits are just one word edits, either inserting or deleting a word.
This indicates that text spans are a good approximation for natural answers, positive news for span-based reading comprehension models.
The remaining one third involve multiple edits.
Although multiple edits are challenging to evaluate using automatic metrics, we observe that many of these answers partially overlap with passage, indicating that word overlap is still a reliable automatic evaluation metric in our setting.
The rest of the answers include counting (5.1\%) and selecting a choice from the question~(1.8\%).

\subsection{Conversation Flow}
\begin{figure}[t]
\centering
\includegraphics[width=\columnwidth]{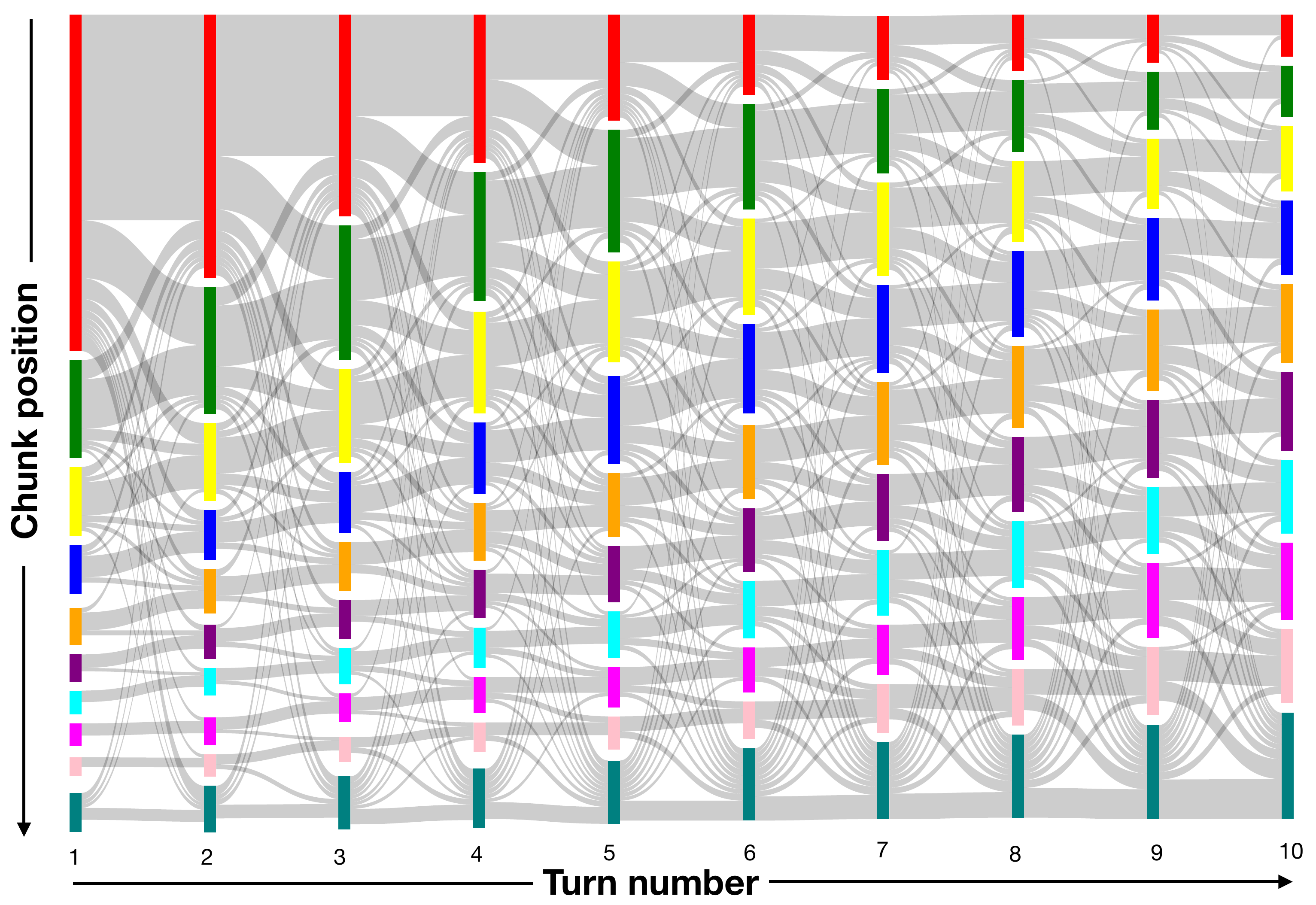}
\caption{Chunks of interest as a conversation progresses.
Each chunk is one tenth of a passage.
The $x$-axis indicates the turn number and the $y$-axis indicates the chunk containing the rationale.
The height of a chunk indicates the concentration of conversation in that chunk.
The width of the bands is proportional to the frequency of transition between chunks from one turn to the next.}
\label{fig:dialog-flow}
\end{figure}

A coherent conversation must have smooth transitions between turns.
We expect the narrative structure of the passage to influence our conversation flow.
We split each passage into 10 uniform chunks, and identify chunks of interest in a given turn and its transition based on rationale spans.
\Cref{fig:dialog-flow} shows the conversation flow of the first 10 turns.
The starting turns tend to focus on the first few chunks and as the conversation advances, the focus shifts to the later chunks.
Moreover, the turn transitions are smooth, with the focus often remaining in the same chunk or moving to a neighboring chunk.
Most frequent transitions happen to the first and the last chunks, and likewise these chunks have diverse outward transitions.


\section{Models}
\label{sec:models}
Given a passage $p$, the conversation history \{$q_1, a_1, \ldots q_{i-1}, a_{i-1}$\} and a question $q_i$, the task is to predict the answer ${a_i}$. Gold answers $a_1$, $a_2$, \ldots, $a_{i-1}$ are used to predict $a_i$, similar to the setup discussed in \Cref{sec:multiple-answers}.

Our task can either be modeled as a conversational response generation problem or a reading comprehension problem.
We evaluate strong baselines from each modeling type and a combination of the two on CoQA.

\subsection{Conversational Models}
Sequence-to-sequence (\textit{seq2seq}) models have shown promising results for generating conversational responses \cite{vinyals_neural_2015,serban_generative_2016,zhang_personalizing_2018}.
Motivated by their success, we use a sequence-to-sequence with attention model for generating answers \cite{bahdanau_neural_2015}.
We append the conversation history and the current question to the passage, as $p\; \mathrm{<}q\mathrm{>}\; q_{i-n} \;\mathrm{<}a\mathrm{>}\; a_{i-n}\; \ldots$ $\mathrm{<}q\mathrm{>}\; q_{i-1} \;\mathrm{<}a\mathrm{>}\; a_{i-1}\;$  $\mathrm{<}q\mathrm{>}\;q_i$, and feed it into a bidirectional LSTM encoder, where $n$ is the size of the history to be used.
We generate the answer using an LSTM decoder which attends to the encoder states. Additionally, as the answer words are likely to appear in the original passage, we employ a copy mechanism in the decoder which allows to (optionally) copy a word from the passage \cite{gu_incorporating_2016,see_get_2017}. This model is referred to as the Pointer-Generator network, PGNet.

\subsection{Reading Comprehension Models}
The state-of-the-art reading comprehension models for extractive question answering focus on finding a span in the passage which matches the question best \cite{seo_bidirectional_2016,chen_reading_2017,yu_fast_2018}.
Since their answers are limited to spans, they cannot handle questions whose answers do not overlap with the passage, e.g., Q$_3$, Q$_4$ and Q$_5$ in \Cref{fig:question-types}.
However this limitation makes them more effective learners than conversational models which have to generate an answer from a large space of pre-defined vocabulary.

We use the Document Reader (DrQA) model of \newcite{chen_reading_2017}, which has demonstrated strong performance on multiple datasets \cite{rajpurkar_squad_2016,labutov_multirelational_2018}.
Since DrQA requires text spans as answers during training, we select the span which has the highest lexical overlap (F1~score) with the original answer as the gold answer.
If the answer appears multiple times in the story we use the rationale to find the correct one.
If any answer word does not appear in the story, we fall back to an additional \textit{unknown} token as the answer (about 17\% in the training set).
We prepend each question with its past questions and answers to account for conversation history, similar to the conversational models.

Considering that a significant portion of answers in our dataset are \textit{yes} or \textit{no} (Table~\ref{tab:squad-coqa-answers}), we also include an augmented reading comprehension model for comparison. We add two additional tokens, \textit{yes} and \textit{no}, to the end of the passage --- if the gold answer is \textit{yes} or \textit{no}, the model is required to predict the corresponding token as the gold span; otherwise it does the same as the previous model.  We refer to this model as Augmented DrQA.



\begin{table*}
\vspace{-1em}
\footnotesize
\centering
\begin{tabular}{l | c c c c c | c c | c | c | c}
&  \multicolumn{5}{|c|}{In-domain} & \multicolumn{2}{c|}{Out-of-dom.} & In-domain & Out-of-dom. & Overall\\
& Child. & Liter. & Mid-High. & News & Wiki. & Reddit & Science &  Overall & Overall &  \\
\midrule
\multicolumn{11}{c}{Development data}\\
\midrule
Seq2seq & 30.6 & 26.7 & 28.3 & 26.3 & 26.1 & N/A & N/A &  27.5 & N/A & 27.5 \\
PGNet & 49.7 & 42.4 & 44.8 & 45.5 & 45.0 & N/A & N/A & 45.4 & N/A & 45.4 \\
DrQA & 52.4 & 52.6 & 51.4 & 56.8 & 60.3 & N/A & N/A &  54.7 & N/A & 54.7 \\
Augmt. DrQA & \bf 67.0 & \bf 63.2 & \bf 63.9 & \bf 69.8 & 72.0 & N/A & N/A & \bf 67.2 & N/A & \bf 67.2  \\
DrQA+PGNet &  64.5 &  62.0 &  63.8 &  68.0 & \bf 72.6 & N/A & N/A &  66.2 & N/A &  66.2 \\
Human & 90.7 & 88.3 & 89.1 & 89.9 & 90.9 & N/A & N/A & 89.8 & N/A & 89.8 \\
\midrule
\multicolumn{11}{c}{Test data}\\
\midrule
Seq2seq & 32.8 & 25.6 & 28.0 & 27.0 & 25.3 & 25.6 & 20.1 & 27.7 & 23.0 & 26.3 \\
PGNet & 49.0 & 43.3 & 47.5 & 47.5 & 45.1 & 38.6 & 38.1 & 46.4 & 38.3 & 44.1 \\
DrQA & 46.7 & 53.9 & 54.1 & 57.8 & 59.4 & 45.0 & 51.0 & 54.5 & 47.9 & 52.6 \\
Augmt. DrQA & \bf 66.0 & 63.3 & 66.2 & \bf 71.0 & 71.3 & 57.7 & 63.0 & \bf 67.6 & 60.2 & \bf 65.4 \\
DrQA+PGNet &  64.2 & \bf 63.7 & \bf  67.1 &  68.3 & \bf 71.4 & \bf 57.8 & \bf 63.1 &  67.0 & \bf 60.4 &  65.1  \\
Human & 90.2 & 88.4 & 89.8 & 88.6 & 89.9 & 86.7 & 88.1 & 89.4 & 87.4 & 88.8 \\
\midrule
\end{tabular}
\caption{Models and human performance (F1 score) on the development and the test data.}
\label{tab:results}
\vspace{-1em}
\end{table*}

\subsection{A Combined Model}
Finally, we propose a model which combines the advantages from both conversational models and extractive reading comprehension models. We use DrQA with PGNet in a combined model, in which DrQA first points to the answer evidence in the text, and PGNet naturalizes the evidence into an answer.
For example, for Q$_5$ in \Cref{fig:question-types}, we expect that DrQA first predicts the rationale R$_5$, and then PGNet generates A$_5$ from R$_5$.

We make a few changes to DrQA and PGNet based on empirical performance.
For DrQA, we require the model to predict the answer directly if the answer is a substring of the rationale, and to predict the rationale otherwise. For PGNet, we provide the current question and DrQA's span predictions as input to the encoder and the decoder aims to predict the final answer.\footnote{We feed DrQA's oracle spans into PGNet during training.}

\section{Evaluation}


\subsection{Evaluation Metric}
Following SQuAD, we use macro-average F1 score of word overlap as our main evaluation metric.\footnote{SQuAD also uses exact-match metric, however we think F1 is more appropriate for our dataset because of the free-form answers.} We use the gold answers of history to predict the next answer.  In SQuAD, for computing a model's performance, each individual prediction is compared against $n$ human answers resulting in $n$~F1 scores, the maximum of which is chosen as the prediction's F1.\footnote{However, for computing human performance, a human prediction is only compared against $n-1$ human answers, resulting in underestimating human performance. We fix this bias by partitioning $n$ human answers into $n$ different sets, each set containing $n-1$ answers, similar to \newcite{choi_quac_2018}.}
For each question, we average out F1 across these $n$ sets, both for humans and models. In our final evaluation, we use $n=4$ human answers for every question (the original answer and 3 additionally collected answers). The articles \textit{a, an} and \textit{the} and punctuations are excluded in evaluation.


\subsection{Experimental Setup}

For all the experiments of seq2seq and PGNet, we use the {OpenNMT} toolkit \cite{klein_opennmt_2017} and its default settings: 2-layers of LSTMs with $500$ hidden units for both the encoder and the decoder. The models are optimized using SGD, with an initial learning rate of $1.0$ and a decay rate of $0.5$. A dropout rate of $0.3$ is applied to all layers.

For the DrQA experiments, we use the implementation from the original paper \cite{chen_reading_2017}. We tune the hyperparameters on the development data: the number of turns to use from the conversation history, the number of layers, number of each hidden units per layer and dropout rate. The best configuration we find is 3 layers of LSTMs with $300$ hidden units for each layer. A dropout rate of $0.4$ is applied to all LSTM layers and a dropout rate of $0.5$ is applied to word embeddings. We used Adam to optimize DrQA models.

We initialized the word projection matrix with GloVe \cite{pennington_glove_2014} for conversational models and fastText \cite{bojanowski_enriching_2017} for reading comprehension models, based on empirical performance.
We update the projection matrix during training in order to learn embeddings for delimiters such as $\mathrm{<}q\mathrm{>}$.

\subsection{Results and Discussion}

\Cref{tab:results} presents the results of the models on the development and test data. Considering the results on the test set, the seq2seq model performs the worst, generating frequently occurring answers irrespective of whether these answers appear in the passage or not, a well known behavior of  conversational models \cite{li_diversitypromoting_2016}.
PGNet alleviates the frequent response problem by focusing on the vocabulary in the passage and it outperforms seq2seq by 17.8 points.
However, it still lags behind DrQA by 8.5 points.
A reason could be that PGNet has to memorize the whole passage before answering a question, a huge overhead which DrQA avoids.
But DrQA fails miserably in answering questions with answers which do not overlap with the passage (see row \textit{No span found} in \Cref{tab:error-analysis}).
The augmented DrQA circumvents this problem with additional yes/no tokens, giving it a boost of 12.8 points.
When DrQA is fed into PGNet, we empower both DrQA and PGNet --- DrQA in producing free-form answers; PGNet in focusing on the rationale instead of the passage.
This combination outperforms vanilla PGNet and DrQA models by 21.0 and 12.5 points respectively, and is competitive with the augmented DrQA (65.1~vs.~65.4).


\paragraph{Models vs. Humans}
The human performance on the test data is 88.8 F1, a strong agreement indicating that the CoQA's questions have concrete answers.
Our best model is 23.4 points behind humans.

\paragraph{In-domain~vs.~Out-of-domain}
All models perform worse on out-of-domain datasets compared to in-domain datasets. The best model drops by 6.6 points. For in-domain results, both the best model and humans find the literature domain harder than the others since literature's vocabulary requires proficiency in English.
For out-of-domain results, the Reddit domain is apparently harder.
While humans achieve high performance on children's stories,  models perform poorly, probably due to the fewer training examples in this domain compared to others.\footnote{We collect children's stories from MCTest which contains only 660 passages in total, of which we use 200 stories for the development and the test sets.}
Both humans and models find Wikipedia easy.

\begin{table}[tp]
\footnotesize
\centering
\begin{adjustwidth}{-1.5em}{}
\begin{tabular}{lp{2.2em}p{2em}p{2em}p{2em}p{2em}p{2em}}
\toprule
Type & Seq2seq & PGNet & DrQA & Augmt.  & DrQA+ & Human\\
 &              &     &          & DrQA  & PGNet & \\
\midrule
\multicolumn{7}{c}{Answer Type} \\
\midrule
Answerable & 27.5 & 45.4 & 54.7 &  67.3 &  66.3 & 89.9 \\
Unanswerable & 33.9 & 38.2 & 55.0 & 49.1  &  51.2 & 72.3 \\
\midrule
Span found & 20.2 & 43.6 & 69.8 &  71.0 &  70.5 & 91.1 \\
No span found & 43.1 & 49.0 & 22.7 &  59.4 & 57.0 & 86.8 \\
\midrule
Named Entity & 21.9 & 43.0 & 72.6 &  73.5 & 72.2 & 92.2 \\
Noun Phrase & 17.2 & 37.2 & 64.9 &  65.3 & 64.1 & 88.6 \\
Yes & 69.6 & 69.9 & 7.9\; &  75.7 & 72.7 & 95.6 \\
No & 60.2 & 60.3 & 18.4 &  59.6 &  58.7 & 95.7 \\
Number & 15.0 & 48.6 & 66.3 & 69.0 &  71.7 & 91.2 \\
Date/Time & 13.7\; & 50.2 & 79.0 &  83.3 & 79.1 & 91.5 \\
Other & 14.1 & 33.7 & 53.5 &  55.6 & 55.2 & 80.8 \\
\midrule
\multicolumn{7}{c}{Question Type} \\
\midrule
Lexical Mat. & 20.7 &  40.7 & 57.2 &  75.5 & 65.7 & 91.7 \\
Paraphrasing &  23.7 & 33.9 & 46.9 & 62.6 &  64.4 & 88.8 \\
Pragmatics  & 33.9 & 43.1 & 57.4 &  64.1 &  60.6 & 84.2 \\
\midrule
No coref. & 16.1  & 31.7 & 54.3 &  70.9 & 58.8 & 90.3  \\
Exp. coref. & 30.4 & 42.3 & 49.0 & 63.4 &  66.7 & 87.1 \\
Imp. coref. & 31.4 & 39.0 & 60.1 &  70.6 & 65.3 & 88.7 \\
\bottomrule
\end{tabular}
\end{adjustwidth}
\caption{Fine-grained results of different question and answer types in the development set. For the question type results, we only analyze 150 questions as described in \Cref{sec:ling-phenomena}.}
\label{tab:error-analysis}
\end{table}

\paragraph{Error Analysis}
\Cref{tab:error-analysis} presents fine-grained results of models and humans on the development set.
We observe that humans have the highest disagreement on the unanswerable questions.
The human agreement on answers which do no overlap with passage is lower than on answers which overlap.
This is expected because our evaluation metric is based on word overlap rather than on the meaning of words.
For the question \textit{did Jenny like her new room?},  human answers \textit{she loved it} and \textit{yes} are both accepted.
Finding the perfect evaluation metric for abstractive responses is still a challenging problem \cite{liu_how_2016,chaganty_price_2018} and beyond the scope of our work.
For our models' performance, seq2seq and PGNet perform well on non-overlapping answers, and DrQA performs well on overlapping answers, due to their respective designs.
The augmented and combined models improve on both categories.


Among the different question types, humans find lexical matches the easiest followed by paraphrasing, and pragmatics the hardest --- this is expected since questions with lexical matches and paraphrasing share some similarity with the passage, thus making them relatively easier to answer than pragmatic questions.
This is also the case with the combined model, but we could not explain the behaviour of other models.
While humans find the questions without coreferences easier than those with coreferences, the models behave sporadically.
Humans find implicit coreferences easier than explicit coreferences.
A conjecture is that implicit coreferences depend directly on the previous turn whereas explicit coreferences may have long distance dependency on the~conversation.

\paragraph{Importance of conversation history}
Finally, we examine how important the conversation history is for the dataset. \Cref{tab:ablations} presents the results with a varied number of previous turns used as conversation history.
All models succeed at leveraging history but the gains are little beyond one previous turn.
As we increase the history size, the performance decreases.

We also perform an experiment on humans to measure the trade-off between their performance and the number of previous turns shown.
Based on the heuristic that short questions likely depend on the conversation history, we sample 300 one or two word questions, and collect answers to these varying the number of previous turns shown.

When we do not show any history, human performance drops to 19.9 F1 as opposed to 86.4 F1 when full history is shown.
When the previous turn (question and answer) is shown, their performance boosts to 79.8 F1, suggesting that the previous turn plays an important role in understanding the current question.
If the last two turns are shown, they reach up to 85.3 F1, almost close to the performance when the full history is shown.
This suggests that most questions in a conversation have a limited dependency within a bound of two turns.


\paragraph{Augmented DrQA vs. Combined Model}
Although the performance of the augmented DrQA is a bit better (0.3~F1 on the testing set) than the combined model, the latter model has the following benefits:
1) The combined model provides a rationale for every answer, which can be used to justify whether the answer is correct or not (e.g., yes/no questions); and
2) we don't have to decide on the set of augmented classes beforehand which helps in answering a wide range of questions like counting and multiple choice (\Cref{tab:aug-vs-comb}).
We also look closer into the outputs of the two models.
Although the combined model is still far from perfect, it does correctly as desired in many examples, e.g., for a counting question, it predicts a rationale \textit{current affairs , politics , and culture} and generates an answer \textit{three}; for a question \textit{With who?}, it predicts a rationale \textit{Mary and her husband , Rick} and then compresses it into \textit{Mary and Rick} for improving the fluency; and for a multiple choice question \textit{Does this help or hurt their memory of the event?} it predicts a rationale \textit{this obsession may prevent their brains from remembering} and answers \textit{hurt}. We think there is still great room for improving the combined model and we leave it to future work.

\begin{table}[tp]
\footnotesize
\centering
\begin{tabular}{cccccc}
\toprule
History & Seq2seq & PGNet & DrQA & Augmt. & DrQA+ \\
size &               &     &            & DrQA &  PGNet \\
\midrule
0 & 24.0 & 41.3 & 50.4 & 62.7 & 61.5 \\
1 & \textbf{27.5} & 43.9 & \textbf{54.7} & 66.8 &  \textbf{66.2} \\
2 & 21.4 & 44.6 & 54.6 & \textbf{67.2} &  66.0 \\
all & 21.0 &  \textbf{45.4} & 52.3 & 64.5 & 64.3 \\
\bottomrule
\end{tabular}
\caption{Results on the development set with different history sizes. History size indicates the number of previous turns prepended to the current question. Each turn contains a question and its answer.}
\label{tab:ablations}
\end{table}

\ignore{
\begin{table}
\footnotesize
\centering
\begin{tabular}{cccccc}
\toprule
history & Seq2seq & PGNet & DrQA & DrQA+ & Human \\
size &               &     &            & PGNet & \\
\midrule
0 & 10.2 & 32.1 & 36.4 & 41.7 & 19.9 \\
1 & 17.8 & 41.8 & 62.8 &  62.9 & 79.8 \\
2 & 12.3 & 39.4 & 65.9 & 63.0 & 85.3 \\
all & 12.7 &  41.4 & 58.5 & 64.1 & 86.6 \\
\bottomrule
\end{tabular}
\caption{Only on \textbf{300} questions. Results on the development set with different history sizes. History size indicates the number of previous turns prepended to the current question. Each turn contains a question and its answer.}
\label{tab:ablations}
\end{table}
}

\begin{table}
\footnotesize
\centering
\begin{tabular}{lccc}
\toprule
&   Augmt. & DrQA+ & Human \\
& DrQA & PGNet \\
\midrule
Yes & \bf 76.2  & 72.5 &  97.7 \\
No & \bf 64.0 & 57.5 &  96.8 \\
Fluency  &  \bf 37.6 & 32.3 &  77.2  \\
Counting  &  8.8 & \bf 24.8  &  88.3 \\
Multiple choice  &  0.0 & \bf 46.4 & 94.3  \\
\bottomrule
\end{tabular}
\caption{Error analysis of questions with answers which do not overlap with the text passage.}
\label{tab:aug-vs-comb}
\end{table}

\ignore{
\begin{table*}
  \center
    \footnotesize
    \begin{tabular}{p{1cm} p{4cm} p{2cm} p{2cm} p{4cm}}
      \toprule
      Category & Question & Augmt. DrQA & \multicolumn{2}{l}{DrQA + PGNet} \\
      & & Answer & Answer & Predicted rationale \\
      \midrule
      Yes & Did Jenny like her new room? & no & \textbf{yes} & She loved her new bedroom \\
      \midrule
      No & Will it be the last? & \textbf{no} & \textbf{no} & definitely won't be the last \\
      \midrule
      Fluency & With who? & Chichi & \textbf{mary and rick} & Mary and her husband , Rick \\
      \midrule
      Counting & How many main topics does the US version cover? & unknown & \textbf{three} & current affairs , politics , and culture \\
      \midrule
      Multiple choice & Does this help or hurt their memory of the event? & unknown & \textbf{hurt} & this obsession may prevent their brains from remembering what actually happened \\
      \bottomrule
    \end{tabular}
    \caption{Example outputs of the combined model (and its predicted rationale) and the Augmented DrQA model. The correct answers are written in bold.}
    \label{tab:example-output}
\end{table*}
}


\section{Related work}
\label{sec:related-work}

We organize CoQA's relation to existing work under the following criteria.

\paragraph{Knowledge source}
We answer questions about text passages --- our knowledge source.
Another common knowledge source is machine-friendly databases which organize world facts in the form of a table or a graph \cite{berant_semantic_2013,pasupat_compositional_2015,bordes_largescale_2015,saha_complex_2018,talmor_web_2018}.
However understanding their structure requires expertise, making it challenging to crowd-source large QA datasets without relying on templates.
Like passages, other human friendly sources are images and videos \cite{antol_vqa_2015,das_visual_2016,hori_endend_2018}.

\paragraph{Naturalness}
There are various ways to curate questions: removing words from a declarative sentence to create a fill-in-the-blank question \cite{hermann_teaching_2015}, using a hand-written grammar to create artificial questions \cite{weston_towards_2015,welbl_constructing_2018}, paraphrasing artificial questions to natural questions \cite{saha_complex_2018,talmor_web_2018} or, in our case, letting humans ask natural questions  \cite{rajpurkar_squad_2016,nguyen_ms_2016}.
While the former enable collecting large and cheap datasets, the latter enable collecting natural questions.

Recent efforts emphasize collecting questions without seeing the knowledge source in order to encourage the independence of question and documents \cite{joshi_triviaqa_2017,dunn_searchqa_2017,kocisky_narrativeqa_2018}.
Since we allow a questioner to see the passage, we incorporate measures to increase independence, although complete independence is not attainable in our setup (\Cref{sec:interface}).
However, an advantage of our setup is that the questioner can validate the answerer on the spot resulting in high agreement data.


\paragraph{Conversational Modeling}
Our focus is on questions that appear in a conversation.
\newcite{iyyer_searchbased_2017} and \newcite{talmor_web_2018} break down a complex question into a series of simple questions mimicking conversational QA.
Our work is closest to \newcite{das_visual_2016} and \newcite{saha_complex_2018} who perform conversational QA on images and a knowledge graph respectively, with the latter focusing on questions obtained by paraphrasing templates.

In parallel to our work, \newcite{choi_quac_2018} also created a dataset of conversations in the form of questions and answers on text passages.
In our interface, we show a passage to both the questioner and the answerer, whereas their interface only shows a title to the questioner and the full passage to the answerer. Since their setup encourages the answerer to reveal more information for the following questions, their average answer length is 15.1 words (our average is 2.7). While the human performance on our test set is 88.8 F1, theirs is 74.6 F1. Moreover, while CoQA's answers can be free-form text, their answers are restricted only to extractive text spans. Our dataset contains passages from seven diverse domains, whereas their dataset is built only from Wikipedia articles about people.

Concurrently, \newcite{saeidi_interpretation_2018} created a conversational QA dataset for regulatory text such as tax and visa regulations. Their answers are limited to \textit{yes} or \textit{no} along with a positive characteristic of permitting to ask clarification questions when a given question cannot be answered.  \newcite{elgohary_dataset_2018} proposed a sequential question answering dataset collected from Quiz Bowl tournaments, where a sequence contains multiple related questions. These questions are related to the same concept while not focusing on the dialogue aspects (e.g., coreference). \newcite{zhou_dataset_2018} is another dialogue dataset based on a single movie-related Wikipedia article, in which two workers are asked to chat about the content. Their dataset is more like chit-chat style conversations while our dataset focuses on multi-turn question answering.\




\paragraph{Reasoning}

Our dataset is a testbed of various reasoning phenomena occurring in the context of a conversation (\Cref{sec:analysis}).
Our work parallels a growing interest in developing datasets that test specific reasoning abilities: algebraic reasoning \cite{clark_elementary_2015}, logical reasoning \cite{weston_towards_2015},  common sense reasoning \cite{ostermann_semeval_2018} and multi-fact reasoning \cite{welbl_constructing_2018,khashabi_looking_2018,talmor_web_2018}.


\paragraph{Recent progress on CoQA}
Since we first released the dataset in August 2018, the progress of developing better models on CoQA has been rapid.
Instead of simply prepending the current question with its previous questions and answers, \newcite{huang_flowqa_2019} proposed a more sophisticated solution to effectively stack single-turn models along the conversational flow.
Others (e.g., \citeauthor{zhu_sdnet_2018},~\citeyear{zhu_sdnet_2018}) attempted to incorporate the most recent pretrained language representation model BERT ~\cite{devlin_bert_2018}\footnote{Pretrained BERT models were released in November 2018, which have demonstrated large improvements across a wide variety of NLP tasks.} into CoQA and demonstrated superior results.
As of the time we finalized the paper (Jan 8, 2019), the state-of-art F1 score on the test set was 82.8.

\section{Conclusions}
In this paper, we introduced CoQA, a large scale dataset for building conversational question answering systems.
Unlike existing reading comprehension datasets, CoQA contains conversational questions, free-form answers along with text spans as rationales, and text passages from seven diverse domains.
We hope this work will stir more research in conversational modeling, a key ingredient for enabling natural human-machine communication.

\newpage

\section*{Acknowledgements}
We would like to thank MTurk workers, especially the Master Chatters and the MTC forum members, for contributing to the creation of CoQA, for giving feedback on various pilot interfaces, and for promoting our hits enthusiastically on various forums.
CoQA has been made possible with financial support from the Facebook ParlAI and the Amazon Research awards, and gift funding from Toyota Research Institute. Danqi is supported by a Facebook PhD fellowship. We also would like to thank the members of the Stanford NLP group for critical feedback on the interface and experiments. We especially thank Drew Arad Hudson for participating in initial discussions, and Matthew Lamm for proof-reading the paper.
We also thank the VQA team and Spandana Gella for their help in generating \Cref{fig:squad-coqa-comp}.

\balance
\bibliography{references}
\bibliographystyle{acl_natbib}



\begin{appendices}
\section*{Appendix}
\subsection*{Worker Selection}
First each worker has to pass a qualification test that assesses their understanding of the guidelines of conversational QA. The success rate for the qualification test is 57\% with 960 attempted workers. The guidelines indicate this is a conversation about a passage in the form of questions and answers, an example conversation and do's and don'ts.
However, we give complete freedom for the workers to judge what is good and bad during the real conversation.
This helped us in curating diverse categories of questions that were not present in the guidelines (e.g., true or false, fill in the blank and time series questions).
We pay workers an hourly wage around 8--15 USD.

\subsection*{Annotation Interface}
\label{sec:annotation-interface}
Figure~\ref{fig:annotation-interface} shows the annotation interfaces for both questioners and answerers.

\subsection*{Additional Examples}
We provide additional examples in \Cref{fig:time-example} and \Cref{fig:no-unknown}.

\begin{figure*}[!htp]
    \centering
    \includegraphics[width=\textwidth]{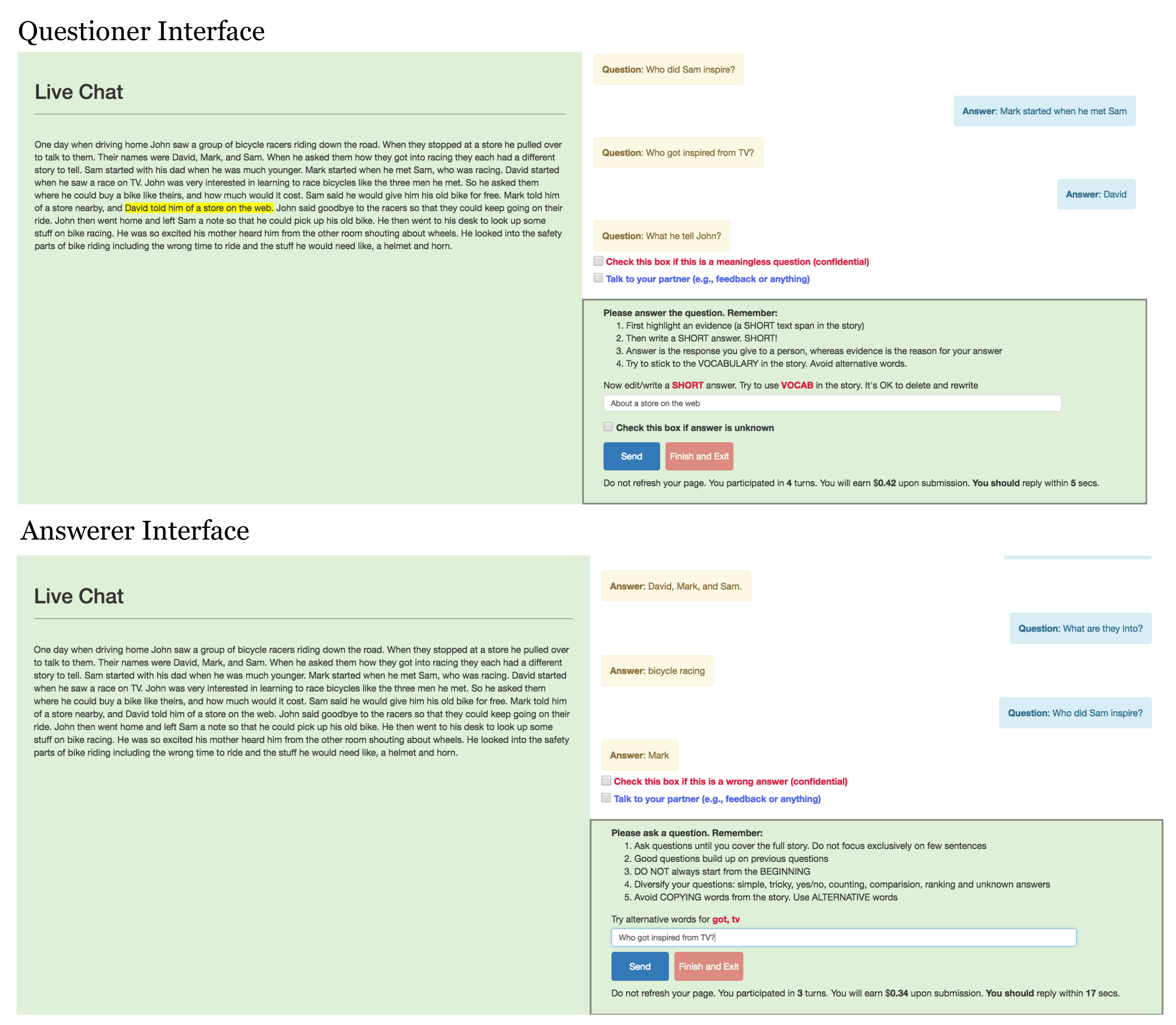}
    \caption{Annotation interfaces for questioner (top) and answerer (bottom).}
    \label{fig:annotation-interface}
    \vspace{10em}
\end{figure*}


\begin{figure}[htp]
  \vspace{2em}
  \footnotesize
  \begin{tabular}{p{\columnwidth}}
  \midrule
  New Jersey is a state in the Northeastern and mid-Atlantic regions of the United States. It is a peninsula, bordered on the north and east by the state of New York; on the east, southeast, and south by the Atlantic Ocean; on the west by the Delaware River and Pennsylvania; and on the southwest by the Delaware Bay and Delaware. New Jersey is the fourth-smallest state by area but the 11th-most populous and the most densely populated of the 50 U.S. states. New Jersey lies entirely within the combined statistical areas of New York City and Philadelphia and is the third-wealthiest state by median household income as of 2016.\\
  \\
  Q:		Where is New jersey located?\\
  A:		In the Northeastern and mid-Atlantic regions of the US. \\
  R: New Jersey is a state in the Northeastern and mid-Atlantic regions of the United States\\
  \\
  Q:		What borders it to the North and East?\\
  A:		New York \\
  R:  bordered on the north and east by the state of New York;\\
  \\
  Q:		Is it an Island?\\
  A:		\textbf{No}. \\
  R: It is a peninsula\\
  \\
  Q:		What borders to the south?\\
  A:		Atlantic Ocean \\
  R:  bordered on the north and east by the state of New York; on the east, southeast, and south by the Atlantic Ocean\\
  \\
  Q:		to the west?\\
  A:		Delaware River and Pennsylvania.\\
  R:  on the west by the Delaware River and Pennsylvania;\\
  \\
  Q:		is it a small state?\\
  A:		Yes. \\
  R:  New Jersey is the fourth-smallest state by area\\
  \\
  Q:		How many people live there?\\
  A:		\textbf{unknown}\\
  R: N/A\\
  \\
  Q:		Do a lot of people live there for its small size?\\
  A:		Yes. \\
  R:  the most densely populated of the 50 U.S. states.\\
  \\
  Q:		Is it a poor state?\\
  A:		\textbf{No}. \\
  R: Philadelphia and is the third-wealthiest state by median household income as of 2016.\\
  \\
  Q:		What country is the state apart of?\\
  A:		United States \\
  R: New Jersey is a state in the Northeastern and mid-Atlantic regions of the United States\\
  \vspace{0em}
  $\ldots$\\
  \bottomrule
  \end{tabular}
  \setcounter{figure}{6}
  \caption{A conversation containing \textit{No} and \textit{unknown} as answers. }
  \label{fig:no-unknown}
  \end{figure}

  \begin{figure}[htp]
  \vspace{2em}
  \footnotesize
  \begin{tabular}{p{\columnwidth}}
  \midrule
  Anthropology is the study of humans and their societies in the past and present. Its main subdivisions are social anthropology and cultural anthropology, which describes the workings of societies around the world, ... Similar organizations in other countries followed: The American Anthropological Association in 1902, the Anthropological Society of Madrid (1865), the Anthropological Society of Vienna (1870), the Italian Society of Anthropology and Ethnology (1871), and many others subsequently. The majority of these were evolutionist. One notable exception was the Berlin Society of Anthropology (1869) founded by Rudolph Virchow, known for his vituperative attacks on the evolutionists. Not religious himself, he insisted that Darwin's conclusions lacked empirical foundation.\\
  \\
  Q:               Who disagreed with Darwin?\\
  A:               Rudolph Virchow \\
  R: Rudolph Virchow, known for his vituperative attacks on the evolutionists. Not religious himself, he insisted that Darwin's conclusions lacked empirical foundation.\\
  \\
  Q:               What did he found?\\
  A:               the Berlin Society of Anthropology \\
  R:  the Berlin Society of Anthropology (1869) founded by Rudolph Virchow\\
  \\
  Q:               In what year?\\
  A:               1869 \\
  R:  the Berlin Society of Anthropology (1869)\\
  \\
  Q:               What was founded in 1865?\\
  A:               the Anthropological Society of Madrid\\
  R:               the Anthropological Society of Madrid (1865) \\
  \\
  Q:               And in 1870?\\
  A:               the Anthropological Society of Vienna\\
  R:  the Anthropological Society of Vienna (1870)\\
  \\
  Q:               How much later was the Italian Sociaty of Anthropology and Ethnology founded?\\
  A:               One year \\
  R: the Anthropological Society of Vienna (1870), the Italian Society of Anthropology and Ethnology (1871)\\
  \\
  Q:               Was the American Anthropological Association founded before or after that?\\
  A:               after \\
  R: The American Anthropological Association in 1902\\
  \\
  Q:               In what year?\\
  A:               1902 \\
  R: The American Anthropological Association in 1902\\
  \\
  Q:               Was it an evolutionist organization?\\
  A:               Yes \\
  R: The majority of these were evolutionist\\
  \vspace{0em}
  $\ldots$\\
  \bottomrule
  \end{tabular}
  \setcounter{figure}{5}
  \caption{In this example, the questioner explores questions related to time. }
  \label{fig:time-example}
  \end{figure}

\end{appendices}

\end{document}